# Invariant 3D Shape Recognition using Predictive Modular Neural Networks

Vasileios Petridis, life member, IEEE

Dept. of Electrical and Computer Engineering, Aristotle University, Thessaloniki, Greece

Email: vpetridi@auth.gr

## Abstract

In this paper PREMONN (PREdictive MOdular Neural Networks) model/architecture is generalized to functions of two variables and to non-Euclidean spaces. It is presented in the context of 3D invariant shape recognition and texture recognition. PREMONN uses local relation, it is modular and exhibits incremental learning. The recognition process can start at any point on a shape or texture, so a reference point is not needed. Its local relation characteristic enables it to recognize shape and texture even in presence of occlusion. The analysis is mainly mathematical. However, we present some experimental results. The methods presented in this paper can be applied to many problems such as gesture recognition, action recognition, dynamic texture recognition etc.

## 1. Introduction

There is a large number of shape descriptors in 3D shape analysis [1-3,13], e.g. curvatures, surface normals, angles, properties of spherical functions, local shape diameters, triangle areas, SIFT and SURF feature descriptors, heat kernel signatures etc. There are 3D shape descriptors that are extracted from the original representation of 3D shapes (point clouds, meshes, implicit functions etc.) and those that use a number of 2D projections (viewbased descriptors). Shape can be represented by a bag of features or a histogram computed out of features [4]. Local descriptors are more robust to occlusion and clutter and better for partial shape retrieval than global descriptors [1]. Deep neural networks were used to detect 3D shape features [1,7]. Also, principal patches [9] have been proposed for invariant shape description [8]. Our approach in this paper exploits the invariance of principal curvatures and certain intrinsic properties of the shape.

PREMONN had been introduced in relation to time series [5,33,34] and applied to problems such as classification of time series, parameter estimation of dynamical systems [5,14,15], prediction problems [5,17], action recognition [16] etc. Time series depend on one variable. In this paper this method is extended to functions that depend on two variables (3-dimensional spaces) and non-Euclidean data. It is presented in the context of 2D and 3D shape and texture recognition. Texture of images is considered a discrete function that depends on two variables in a 3D Euclidean space. Also, recognition of surfaces in a 3D Euclidean space is formulated as function recognition in a non-Euclidean space and applied to 3D invariant shape recognition.

PREMONN relies on local relation, it is modular and exhibits incremental learning. The recognition process can start at any point on a curve or surface or texture, so a reference point is not needed. In this way calculation of correspondences is not required. Its local relation characteristic enables it to recognize shape and texture more robustly in presence of occlusion or clutter. Any small patch of the surface is sufficient for recognition. Also, it is suitable for partial 3D shape retrieval.

In section 2 PREMONN model/architecture is reviewed. In section 3 the method is applied to curves. In section 4 the method is generalized to functions of two variables and to non-Euclidean spaces. It is applied to 3D invariant shape recognition and texture recognition [24,26]. Implementation issues and some experimental results are presented in section 5. More extensive results will be presented in a future paper.

## 2. Review of PREMONN

Suppose N time series $y^n(t_1)$, $y^n(t_2)$, … are generated by N unknown source functions $F_n$, n = 1, 2, ..., N according to the following equation ($y_i$=y($t_i$)),

$$y^n_i = F_n(y^n_{i-1}, y^n_{i-2}, …, y^n_{i-M})+noise \tag{1}$$

The noise process may be of unknown characteristics. When for t=1,2,3,… a time series y(1), y(2), … is observed, generated by one of the N sources, the time series classification task is to identify the source that generates the time series using the observations y(1), y(2), …. To this end N neural network predictors, $f_n(.)$, (for n=1,2,…,N) are trained offline, one for each source. The nth time series (generated by $F_n$) from the training set is used to train offline a neural network predictor, $f_n(.)$, which approximates $F_n$. During training the inputs to the neural network are M past observations $y^n_{i-1}$, $y^n_{i-2}$, …, $y^n_{i-M}$ of the nth time series and its output is the estimate $f_n(y^n_{i-1}, y^n_{i-2}, …, y^n_{i-M})$. In the online recognition phase a test time series $y_1$, $y_{2,…}$is presented and the predictions, $\hat{y}^n_i$ = $f_n(y_{i-1}, y_{i-2}, …, y_{i-M})$ (of all N predictors) and prediction errors $e^n_i$=$y_i$-$\hat{y}^n_i$ , n=1,2,…,N are calculated. Then the credit functions $p^n(.)$, n=1,2,…,N, corresponding to the N predictors, are calculated on the basis of prediction errors. A high credit value means that the respective source has a high probability that has generated the test time series.

The algorithm for the recursive online computation of the credit functions is known as PREdictive MOdular Neural Network (PREMONN) classification algorithm and is implemented by the parallel operation on N predictive neural modules. It has been applied to many problems that can be formulated as time series recognition ones, such as classification of visually evoked responses used for diagnosing neuroophthalmological disorders, prediction of short-term electric loads, parameter estimation of dynamical systems, action recognition etc. [5, 14-17].

## Basic PREMONN Classification Algorithm

### Training phase

N neural network predictors $f_n(.)$ (for n=1,2,…,N) are trained offline. At t=0, N arbitrary initial values $p_0^n$ are chosen which satisfy

$$0 < p_0^n < 1, \qquad \sum_{n=1}^{N} p_0^n = 1 \qquad (2)$$

### Main online phase

For time instant i=1,2,…

For n=1,2,…,N compute

Predictions:

$$\hat{y}_i^n = f_n(y_{i-1}, y_{i-2}, \ldots, y_{i-M}) \qquad (3)$$

Prediction errors:

$$e_i^n = y_i - \hat{y}_i^n \qquad (4)$$

Credit functions:

$$p_i^n = \frac{p_{i-1}^n \, e^{\frac{(e_i^n)^2}{2\sigma^2}}}{\sum_{m=1}^{N} p_{i-1}^m \, e^{\frac{(e_i^m)^2}{2\sigma^2}}} \qquad (5)$$

Next n.

At instant i the time series is classified to the source n which maximizes $p_i^n$.

Next i.

We can say that the point $t_i$ constitutes the output domain $D_{out} = \{t_i\}$ and the points $t_{i-1}, t_{i-2}, \ldots, t_{i-M}$ constitute the input domain $D_{in} = \{t_{i-1}, t_{i-2}, \ldots, t_{i-M}\}$. Therefore we can write

$$\hat{y}_{\text{Dout}}^n = \hat{y}_i^n \qquad (6)$$

$$y_{\text{Din}} = [y_{i\text{-}1}, y_{i\text{-}2}, \ldots, y_{i\text{-}M}] \qquad (7)$$

Hence, equ. (3) can be written,

$$\hat{y}_{\text{Dout}}^n = f_n(y_{\text{Din}}) \qquad (8)$$

Predictors $f_n(.)$ are NAR (Nonlinear AutoRegressive) models [10] of dynamical systems. The neural networks predictors need not be very accurate. PREMONN works as long as the right model produces prediction errors that are smaller than the ones produced by all other predictors.

This algorithm can be used also in cases the time series exhibits source switching, that is the time series is not generated by a single source. We have assumed that the predictor functions are neural networks but any other predictor function can be used. In fact, different predictor functions can be used in the same implementation. Also, different credit functions can be used. In other words, there can be many variants of the PREMONN algorithm. PREMONN algorithm is modular, therefore exhibits parallelism. Also, training time scales linearly with the number of sources i.e. classes of the classification task.

A small value of the parameter $\sigma$ speeds the algorithm up, as far as convergence and switching is concerned, but makes the algorithm more sensitive to noise fluctuations. A large value of $\sigma$ makes the algorithm less sensitive to noise fluctuations but slows the algorithm down.

## 3. Plane and Space Curves

The same method can be applied to the problem of plane and space curves recognition. In this case variable t is replaced by s which is the actual length of the curve. For invariant shape (curve) recognition we must use invariant quantities.

It is well known that the curvature of plane curves, $\kappa$, is an invariant. Also, a plane curve C is uniquely determined (except for translation and rotation) if the function of curvature with respect to s, $\kappa(s)$, is specified [11]. Hence, since $\kappa(s)$ is invariant, the shape of C is uniquely determined (even in case C is subject to translation and rotation) when $\kappa(s)$ is specified.

In space curves the curvature, $\kappa$, is an invariant. Torsion, $\tau$, is also an invariant. A 3D curve C is uniquely determined (except for translation and rotation) if the functions of curvature and torsion, $\kappa(s)$ and $\tau(s)$, are specified [11]. Hence, since $\kappa(s)$ and $\tau(s)$ are invariants, the shape of C is uniquely determined (even in case C is subject to translation and rotation) when $\kappa(s)$ and $\tau(s)$ are specified.

In general coordinates $x^i$, i=1,2,3 curvature $\kappa(s)$ is the magnitude of the vector which is the intrinsic derivative of the tangent vector $T^i$, [11]

$$\kappa(s) = \left| \frac{\delta T^i}{\delta s} \right| \tag{9}$$

In orthogonal cartesian coordinates,

$$\kappa(s)^2 = \left( \frac{d^2 x^1}{ds^2} \right)^2 + \left( \frac{d^2 x^2}{ds^2} \right)^2 + \left( \frac{d^2 x^3}{ds^2} \right)^2 \tag{10}$$

We restrict ourselves to the case of plane curves in which case $\tau(s)=0$. Suppose we would like to identify a shape in a picture. The goal is the calculation of curvature when the curve is given as a set of points [12]. The shape-curve can be represented by two functions x=x(s) and y=y(s) where x and y are the pixel coordinates. Here, we use a very simple technique. The length, $\Delta s$, of a section of the curve between two curve points is calculated easily by tracing the pixels of the curve between these two points. If the tracing moves from one pixel-point to its next along x or y we consider that the length of the curve between these two consecutive points is one unit of length along the curve. If it moves to a diagonal pixel the length between these two points is $\sqrt{2}$ (in case of square pixels). In this way a series of values $(x_i, s_i)$ and $(y_i, s_i)$ i=1, 2, 3, .... is generated which is used to calculate an approximation x=x(s) and y=y(s) of the curve. In our experiments we used a polynomial approximation. Then the curvature is calculated at each point by means of equ.10 for plane curves. In this way PREMONN classification algorithm remains the same as in the case of time series, the only difference being that parameter t is replaced by parameter s.

In case the curve is not complex a single network can be trained on it. If it is complex the curve can be divided into subregions and a separate network is trained on each one of them. In this case the shape can be represented as a bag of networks or as a histogram or as a string of networks.

## 4. Texture and 3D Surfaces

Generalizing the PREMONN approach of the previous paragraphs we can consider source functions that depend on two variables, x and y. These functions are defined on a mesh constituted by the points $(x_1,y_1)$, $(x_1,y_2)$, ... , $(x_2,y_1)$, $(x_2,y_2)$, ... , $(x_3,y_1)$, $(x_3,y_2)$,.... etc. In this case we have a two-dimensional discrete function indexed by two indices, i and j (two dimensional case z=z(x,y)) unlike the case of time series where we have a discrete function indexed by one index i (one dimensional case y=y(t)). In case the function z=z(x,y) is defined on a plane the mesh is orthogonal cartesian. In general, the mesh need not be orthogonal cartesian as is the case in the problem of 3D surface recognition as we shall see below.

We present first the case that z=z(x,y) is defined on a plane and the mesh is orthogonal cartesian. We consider first the problem of texture recognition as an intermediate step towards 3D surface recognition.

**Texture**

Texture of a gray scale image is considered as a discrete function of two variables. The two independent variables define the position of a pixel and the dependent variable is the intensity of this pixel. Therefore, the mesh on which the function is defined is orthogonal cartesian. The problem of recognizing one of N textures can be formulated as a discrete function recognition problem by assuming that the observed discrete function, $z_{ij}^n$, is generated by one of N sources, each source corresponding to one of the N textures. The classification task consists in estimating which source generates the observed discrete function.

The N unknown source functions $F_n$, n = 1, 2, . .., N generate N functions according to the following equation,

$$z_{ij}^n = F_n(z_{i,j-1}^n, z_{i,j-2}^n, \dots, z_{i,j-L}^n, z_{i-1,j}^n, z_{i-1,j-1}^n, \dots, z_{i-1,j-L}^n, \dots, z_{i-M,j}^n, z_{i-M,j-1}^n, \dots z_{i-M,j-L}^n)$$ (11)

where

$$z_{ij}^n = z^n(x_i, y_j)$$ (12)

As in section 2, we can write

$$D_{out} = \{(x_i, y_j)\}$$ (13)

and

$$D_{in} = \{(x_i, y_{j-1}), (x_i, y_{j-2}), \dots, (x_i, y_{j-L}), (x_{i-1}, y_j), (x_{i-1}, y_{j-1}), \dots, (x_{i-1}, y_{j-L}), \dots, (x_{i-M}, y_j), (x_{i-M}, y_{j-1}), \dots,$$

$$(x_{i-M}, y_{j-L})\}$$ (14)

$D_{out}$ may include more than one point as well. It should be stressed that $D_{in}$ and $D_{out}$ can be defined in many ways. Examples are shown in fig. 1.

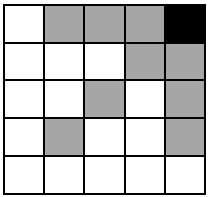 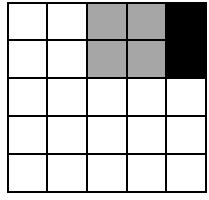 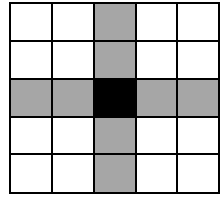 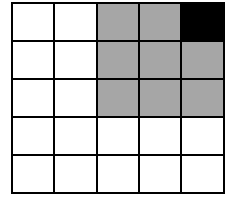

Fig.1a          Fig.1b          Fig.1c          Fig.1d

The gray cells (each cell corresponds to a mesh point) define the input domain though the black ones define the output domain.

In this paper we assume, without loss of generality, that $D_{out}$ consists of one mesh point.

For short equ.(11) is written

$$z^n_{Dout}=F_n(z^n_{Din}) \tag{15}$$

The union of $D_{in}$ and $D_{out}$ (in the same relative position) define the input-output domain D.

For each of the N different classes of textures a sample function is given for training purposes. This is the training set. The sample function in the nth class of the training set is supposed to be generated by the nth source and is used to train offline a neural network predictor, $f_n(..)$, which approximates $F_n$. During training the inputs to the neural network are the values of the function on $D_{in}$ and the required output is the value of the function on $D_{out}$.

In the online recognition phase, input-output domain D is positioned in a randomly chosen position in the sample of texture to be classified. Then D moves around so that the whole area of texture is scanned. During scanning, each position of D constitutes a scanning step. At each scanning step, which is indexed by $l$, predictions of the n predictors,

$$\hat{z}^{nl} = \hat{z}^{nl}_{Dout}=f_n(z^l_{Din}) \tag{16}$$

and prediction errors

$$e^n_l=z^l - \hat{z}^{nl}, \quad \text{n=1, 2, ..., N} \tag{17}$$

are calculated (for $l = 1,2, ....$). $\hat{z}^{nl}$ is, at the $l$ scanning step, the output of the nth predictor on $D_{out}$ when the input to the predictor are the observed values of z on $D_{in}$, $z^l_{Din}$, (the position of $D_{out}$ and $D_{in}$ is the one corresponding to the $l$ scanning step). $z^l$ is the observed value of the discrete function (texture to be classified) at $D_{out}$ at the $l$ scanning step. In case of time series scanning steps are indexed by i.

In case of a color image three discrete functions are defined on the mesh, one for each color. Therefore, we have three functions, $z^{pn}_{ij} = z^{pn}(x_i, y_j)$, p=1,2,3. Equ. (15) is written in this case

$$z^{pn}_{Dout}=F^p_n(z^{pn}_{Din}) \quad \text{p=1,2,3} \tag{18}$$

Three predictors, $f^p_n(..)$ p=1,2,3 are trained on each sample texture, one for each color.

At each scanning step, predictions

$$\hat{z}^{pnl} = \hat{z}^{pnl}_{Dout}=f^p_n(z^{pl}_{Din}) \tag{19}$$

are calculated. Using vector notation,

$$\hat{Z}^{nl}=\begin{bmatrix} \hat{z}^{1nl} \\ \hat{z}^{2nl} \\ \hat{z}^{3nl} \end{bmatrix} \tag{20}$$

The prediction error is

$$e_l^n = \left| Z^l - \hat{Z}^{nl} \right| \qquad n=1, 2, \ldots, N \tag{21}$$

$$Z^l = \begin{bmatrix} z^{1l} \\ z^{2l} \\ z^{3l} \end{bmatrix} \tag{22}$$

where $Z^l$ is the observed discrete vector function at $D_{out}$ at scanning step $l$.

The prediction error is the Euclidean distance between $Z^l$ and $\hat{Z}^{nl}$. The sequence of scanning steps generates a sequence of predictions and prediction errors indexed by $l$. The credit functions $p_l^n(.)$, n=1, 2, …, N, are calculated on the basis of prediction errors. A high credit value means that the respective source has a high probability that has generated the observed function.

We repeat the PREMMON classification algorithm for this case:

## Offline phase

Neural network predictors $f_n^p(.)$ (p=1,2,3 and n=1, 2, …, N) are trained offline. At $l$=0 N arbitrary initial values $p_0^n$ are chosen which satisfy

$$0 < p_0^n < 1, \qquad \sum_{n=1}^{N} p_0^n = 1 \tag{23}$$

## Main online phase

For $l$ =1, 2, …

For n=1, 2, …, N compute

Predictions:

$$\hat{z}^{pnl} = \hat{z}_{Dout}^{pnl} = f_n^p(z_{Din}^{pl}) \qquad p=1,2,3 \tag{24}$$

$$\hat{Z}^{nl} = \begin{bmatrix} \hat{z}^{1nl} \\ \hat{z}^{2nl} \\ \hat{z}^{3nl} \end{bmatrix} \tag{25}$$

Prediction errors:

$$e_l^n = \left| Z^l - \hat{Z}^{nl} \right| \qquad n=1, 2, \ldots, N \tag{26}$$

Credit functions:

$$p_l^n = \frac{p_{l-1}^n \; e^{-\frac{(e_l^n)^2}{2\sigma^2}}}{\sum_{m=1}^N p_{l-1}^m \; e^{-\frac{(e_l^m)^2}{2\sigma^2}}} \tag{27}$$

Next n.

At scanning step $l$ the function is classified to the source n which maximizes $p_l^n$.

Next $l$.

The mesh in the offline training phase should be the same as in the online recognition phase.

In this case predictors $f_n^p(.)$ are equations in two independent variables and can be thought of as dynamical systems. Dynamical systems were used for dynamic texture categorization [30] as well.

**Shape**

Surfaces in three-dimensional Euclidean space, $R^3$, are two-dimensional manifolds embedded into $R^3$. They are colloquially called 3D shapes.

A surface S is represented by three equations $x^1 = x^1(u^1, u^2)$, $x^2 = x^2(u^1, u^2)$, $x^3 = x^3(u^1, u^2)$ where $u^1$ and $u^2$ are the curvilinear coordinates. These equations are usually written $x^i = x^i(u^\alpha)$, i=1, 2, 3 and $\alpha$=1, 2. Setting $u^1$=constant the above equations define a curve lying on the surface S which is called the $u^2$-curve. Similarly setting $u^2$=constant we obtain the $u^1$-curve.

$$a_{\alpha\beta}, \quad \alpha\text{=1,2 and }\beta\text{=1,2} \quad \text{ i.e. } a_{\alpha\beta} = \begin{bmatrix} a_{11} & a_{12} \\ a_{21} & a_{22} \end{bmatrix} \tag{28}$$

is the symmetric covariant metric tensor of the first fundamental quadratic form and

$$b_{\alpha\beta}, \quad \alpha\text{=1,2 and }\beta\text{=1,2} \quad \text{ i.e. } b_{\alpha\beta} = \begin{bmatrix} b_{11} & b_{12} \\ b_{21} & b_{22} \end{bmatrix} \tag{29}$$

is the symmetric surface tensor of the second fundamental quadratic form of the surface. The roots of (solving for $\kappa$)

$$\left| b_{\alpha\beta} - \kappa a_{\alpha\beta} \right| = 0 \quad \alpha\text{=1,2 and } \beta\text{=1,2} \tag{30}$$

are the principal curvatures $\kappa_1$ and $\kappa_2$ at the given point of the surface.

For $\kappa_1$ and $\kappa_2$ the corresponding principal directions, on the surface at the given point, are

$\lambda_1^\beta = \begin{bmatrix} \lambda_1^1 \\ \lambda_1^2 \end{bmatrix}$ and $\lambda_2^\beta = \begin{bmatrix} \lambda_2^1 \\ \lambda_2^2 \end{bmatrix}$ respectively, determined by [11],

$$\left(b_{\alpha\beta} - \kappa_1 a_{\alpha\beta}\right)\lambda_1^{\beta} = 0 \tag{31a}$$

$$\left(b_{\alpha\beta} - \kappa_2 a_{\alpha\beta}\right)\lambda_2^{\beta} = 0 \tag{31b}$$

If the space coordinates are orthogonal cartesian ($x^1$=x, $x^2$=y, $x^3$=z) and the surface coordinates are $u^1$=x and $u^2$=y the vectors $\lambda_1^{\beta}$ and $\lambda_2^{\beta}$ referred to the space coordinates are given by

$$g_1^q = \begin{bmatrix} \lambda_1^1 \\ \lambda_1^2 \\ \lambda_1^1 \frac{\partial z}{\partial x} + \lambda_1^2 \frac{\partial z}{\partial y} \end{bmatrix} \qquad q=1,2,3 \tag{32a}$$

$$g_2^q = \begin{bmatrix} \lambda_2^1 \\ \lambda_2^2 \\ \lambda_2^1 \frac{\partial z}{\partial x} + \lambda_2^2 \frac{\partial z}{\partial y} \end{bmatrix} \qquad q=1,2,3 \tag{32b}$$

For invariant (with respect to translation and rotation) recognition of a 3D surface we use the two principal curvatures, $\kappa_1$ and $\kappa_2$ (which are invariant and determine the surface), defined on a mesh that lies on the net of lines of curvature of the surface. In this case the surface is represented by the two functions $\kappa_1(u^1, u^2)$ and $\kappa_2(u^1, u^2)$ defined on a mesh that lies on the lines of curvature of the surface. That is, the net of lines of curvature are the coordinate curves and $u^1$ and $u^2$ are the curvilinear coordinates along the lines of curvature. Two predictors, $f_n^p(.)$ p=1,2 are trained on each training sample, one for $\kappa_1$ and another for $\kappa_2$. In this case the prediction at scanning step $l$ of the nth predictor is written

$$\widehat{K}^{nl} = \begin{bmatrix} \hat{\kappa}_1^{nl} \\ \hat{\kappa}_2^{nl} \end{bmatrix} \tag{33}$$

The prediction error at scanning step $l$ of the nth predictor is,

$$e_l^n = \left| K^l - \widehat{K}^{nl} \right| \quad \text{n=1, 2, ..., N} \tag{34}$$

PREMONN algorithm and the definition of input and output domains for shape recognition are the same as the ones for texture recognition apart from the calculation of prediction error. Simply, equations (25) and (26) are substituted by (33) and (34) respectively. Similarly, the Gauss and mean curvature can be used which are invariant as well.

In this case that the net of lines of curvature is the coordinate net the resulting mesh lies on the lines of curvature which although they form an orthogonal net they are not plane curves in general.

Shape can be represented by a point cloud or any other kind of representation. For the calculation of principal curvatures $\kappa_1$ and $\kappa_2$ (using equ.30) at a given point P on the surface the tensor components $a_{11}$, $a_{12}$, $a_{22}$, $b_{11}$, $b_{12}$ and $b_{22}$, which are functions of the derivatives of the surface equations, must be calculated. There is extensive literature on calculating the curvature and other quantities from point clouds [18,19, 27-29]. A simple approach is the calculation of an approximation of the surface around point P, for example a polynomial equation z=g(x,y) which approximates the surface around P. Then using the derivatives of z=g(x,y) the tensors' components can be calculated. The mesh around P that lies on the lines of curvature can be determined using the principal directions. Starting at P (which by assumption is the point with $u^1$=0 and $u^2$=0, for short P(0,0)) the mesh point, P(1,0), will be at a distance $\Delta s$ from P(0,0) along the $u^1$-line through P(0,0) in the direction $\lambda_1^\beta$ (Fig. 2). Also, point P(0,1) will be at a distance $\Delta s$ from P(0,0) along the $u^2$-line through P(0,0) in the direction $\lambda_2^\beta$. P(2,0), P(3,0), P(-1,0), P(-2,0) etc and P(0,2), P(0,3), P(0,-1), P(0,-2) etc can be calculated in a similar way. P(1,1) is the point on S at the intersection of two curves: the $u^2$-line through P(1,0) in the direction $\lambda_2^\beta$ and the $u^1$-line through P(0,1) in the direction $\lambda_1^\beta$. Mesh points P(1,2), P(2,2), P(1,-1), P(-1,-1), P(-1,2) etc can be determined in the same way. Such an approach can work in case the surface is not complex and the level of noise is not too high. Alternatively, methods like the one presented in [18] can be used for the determination of the lines of curvature.

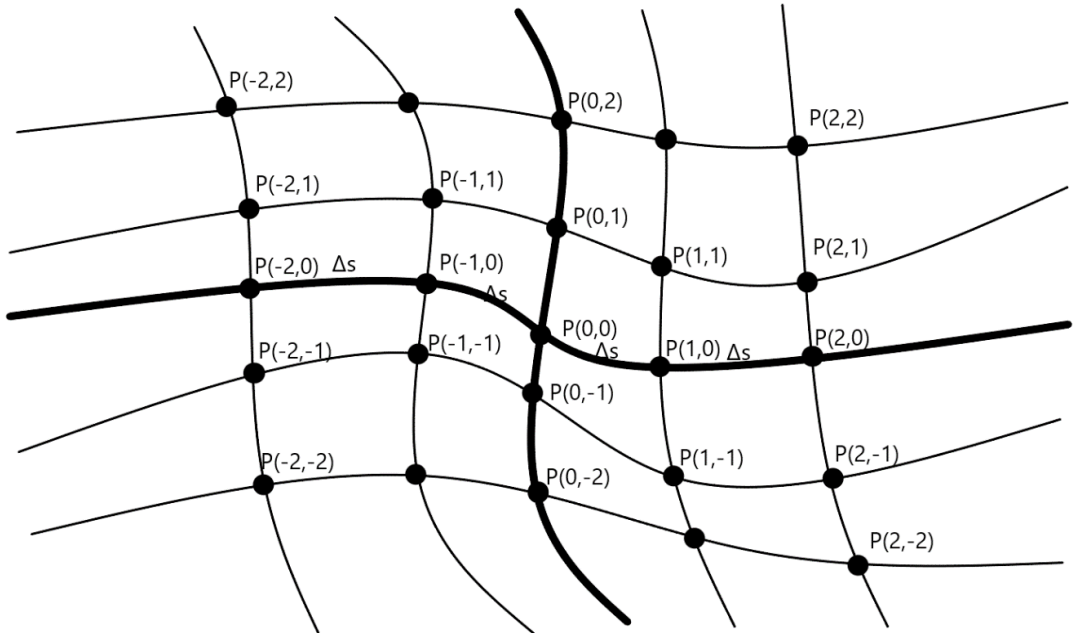

Fig.2   Mesh on line of curvatures

No registration or correspondence problem arises because all the PREMONN algorithm does is to determine which local relation (of the N local relations defined by the N predictors) is "closest" (in the sense that exhibits the smallest prediction error) to the observed one. In a way the local relation (which can be viewed as a local constraint) which is learned by a neural network "characterizes" the shape.

In case the surface is not complex a mesh that covers the whole surface can be generated using the method described above. If the surface is rather complex we can either use methods like in [18] to generate a mesh that covers the whole surface or segment (in certain cases surfaces are composed of patches anyway [6, 20-22]) the surface into subregions (or patches) and generate a mesh on each subregion. In the first case a single neural network is trained on the surface. In the second case a neural network is trained on each subregion and the surface can be represented as a bag of neural networks or a histogram of neural networks or a graph, the nodes of which are the neural networks. Implicit functions can be used as well [31]. Convolutional networks have been used for shape segmentation as well [23].

## 5. Implementation and Experimental Results

We give here, a brief description of some results in [25]. The dataset from Columbia University Image Library (COIL-20) [35] was used. The dataset consists of 20 different objects in 72 positions each, with one picture for each position. Each position results from object rotation around its vertical axis. They are grayscale images with 128x128 pixels resolution. Ten objects were used in the experiments. For each object 40 pictures were used which were grouped into 8 poses: front, rear, left, right, rear-left, rear-right, front-left, front-right. Each pose consisted of 5 pictures. 4 pictures of each pose were used for training and 1 for testing. Therefore, for each object there were 32 positions for training and 8 for testing, resulting in 320 positions for training and 80 for testing in total.

For each picture the contour's equations in parametric form x=x(s) and y=y(s) were estimated (s is the curve's length starting from an arbitrary point on the contour) and the curvature was calculated at each point by means of equ.10 for plane curves as presented in section 3. For the estimation of the previous equations $2^{nd}$ order polynomials turned out to be adequate. In this experiment a filter was used to filter out the curvature's noise. The input to the neural networks was the filtered curvature. Small neural networks were used that can be trained with few training data.

For each one of the 320 contours to be used for training, a sequence of values of $\kappa(s)$ was generated, $\kappa(s_1)$, $\kappa(s_2)$, …: the training sequence. Then, starting at an arbitrary term of the sequence corresponding to a point on the contour (the starting point), each one of the 320 training sequences was divided into a number of subsequences (corresponding to subregions of the contour) using unsupervised learning.  On each subregion a neural network was trained (as explained in sections 2 and 3). In this way for each picture's contour a number of neural networks forming a string was generated. Two approaches were tried. A picture's contour was represented either as a histogram of networks generated by the above process or as a string of networks.

The objective was pose classification. During testing the sequence of values of $\kappa(s)$ along the testing picture's contour was presented to all neural networks. The network with the smallest prediction error within a part of the sequence was the winner i.e. characterized the contour area that corresponded to this part of the sequence. Different networks were winners within different parts of the sequence forming a string of networks. In this way the testing contour was represented either as a histogram of networks in it or as a string of networks.

In the first approach the histogram was used. For each pose the histogram (which was the mean histogram of the four picture's contour, used for training) was computed which was taken to represent the pose. In this way 80 histograms were generated. During testing the histogram of the testing picture's contour was computed in the way explained above. This was compared with the 80 pose histograms and the testing picture was classified to the pose, the histogram of which was closest to the testing histogram. A number of experiments were performed using different neural networks and various values of the prediction error threshold (for unsupervised learning). When in testing the contour was traced starting from the same starting point as the one in training the success rate was between 81.25% (65/80 pictures) and 90% (72/80 pictures). When the starting point was different between training and testing the success rate was between 80% (64/80) and 88.75% (71/80). Also, experiments were performed using all 720 pictures [25]. The success rate was between 83.19% (599/720) and 85.97% (619/720).

In the second approach the string of networks was used. In the experiments described above 4 strings for each pose resulted from the 4 pictures for training and one from the testing picture. That is 4 strings /pose*8 poses*10 objects=320 resulted from the training pictures and 80 from the testing ones. During testing the string of the testing picture was computed and compared with the strings obtained from the training pictures. Levenstein distance was used. Since the resulting strings depend on the starting point of tracing the picture's contour, all possible strings resulting from string shifts were computed. In this case success rate was raised to 79/80 and in some cases to 80/80. The string of networks captured spatial relation, therefore resulted in better success rates than those of histograms approach.

## 6. Conclusions and Future Work

In this paper PREMONN is extended to functions that depend on two variables. In a similar way it can be extended to functions of more variables. Texture recognition is formulated as function recognition in a 3D Euclidean space. Also, invariant 3D surface recognition is formulated as function recognition in a non-Euclidean space.

The method presented here relies on local relation, it is modular and exhibits incremental learning. The recognition process can start at any point on a curve or surface or texture. Therefore, there is no need for calculation of a reference point or correspondences. Its local relation characteristic creates the possibility that this method recognizes shape and texture more robustly in the presence of occlusion or clutter. Also, it is suitable for partial 3D shape retrieval and segmentation of texture or shape. If the shape (curve or surface) can be represented by a single network, then even a small region is sufficient to identify the whole shape (i.e. the algorithm scans a small region of the shape).

In case of complex surfaces we can either use methods like in [18] to generate a mesh that covers the whole surface or segment the surface into patches and generate a mesh on each patch. In the first case a single neural network will be trained on the surface. In the second case a neural network will be trained on each patch and the surface will be represented as a bag of neural networks or a histogram of neural networks or a graph the nodes of which are the neural networks.

Modularity of the method enables the parallel implementation of the algorithm. This feature is very useful especially in cases of problems with large number of networks. Incremental learning is a strong characteristic of the method enabling it to incorporate new input data and to be applied to big data problems.

In the future HPREMONN (Hierarchical PREdictive MOdular Neural Networks), which is a modification of PREMONN in a way that can be applied to problems with a very large number of networks [32], will be used for classification of shapes and texture of large data bases.

Classification" World Congress on Computational Intelligence- Proc. Int. Joint Conference on Neural Networks, (WCCI20010-IJCNN010), pp. 179-186, Barcelona, Spain, 18-23 July 2010.